\documentclass[]{amsart}
\usepackage{graphicx}
\usepackage{amssymb}
\usepackage{amsmath}

\usepackage{algorithm}
\usepackage{algorithmic}
\usepackage{subfigure}
\usepackage{color}
\usepackage{multirow}
\usepackage{wasysym}

\begin{document}

%\title{EDA using RBM: Exploiting the real power of parallel computer architectures}
\title[Denoising Autoencoders for fast Combinatorial BBO]{Denoising Autoencoders for fast Combinatorial Black Box Optimization}
%\numberofauthors{1} %  in this sample file, there are a *total*
%1st. author
\author[]{Malte Probst}
\address[]{Johannes Gutenberg-Universit\"at Mainz\\
Dept.~of Information Systems and Business Administration\\
Jakob-Welder-Weg 9, 55128 Mainz, Germany}
\email{probst@uni-mainz.de}
\urladdr{http://wi.bwl.uni-mainz.de}
\date{20 January 2015}

\begin{abstract}
Estimation of Distribution Algorithms (EDAs) require flexible probability models that can be efficiently learned and sampled. Autoencoders (AE) are generative stochastic networks with these desired properties. We integrate a special type of AE, the Denoising Autoencoder (DAE), into an EDA and evaluate the performance of DAE-EDA on several combinatorial optimization problems with a single objective. We asses the number of fitness evaluations as well as the required CPU times. We compare the results to the performance to the Bayesian Optimization Algorithm (BOA) and RBM-EDA, another EDA which is based on a generative neural network which has proven competitive with BOA. For the considered problem instances, DAE-EDA is considerably faster than BOA and RBM-EDA, sometimes by orders of magnitude. The number of fitness evaluations is higher than for BOA, but competitive with RBM-EDA. These results show that DAEs can be useful tools for problems with low but non-negligible fitness evaluation costs.
\end{abstract}
\maketitle
%==============================================================================
%------------------------------------- INTRO ----------------------------------
%==============================================================================
\section{Introduction}
\label{intro}
Estimation of Distribution Algorithms (EDA, \cite{Muehlenbein1996,larranaga2002estimation}) are metaheuristics for combinatorial and continuous non-linear optimization. They maintain a population of solutions which they improve over consecutive generations. They estimate how likely it is that decisions are part of an optimal solution, and try to uncover the dependency structure between the decision variables. 
This information is obtained from the population by the estimation of a probabilistic model. If a model generalizes the population well, random samples drawn from the model have a structure and solution quality that is similar to the population itself. Repeated model estimation, sampling, and selection steps can solve difficult optimization problems.
Simple models, such as factorizations of univariate frequencies, can be quickly estimated from a population, but they cannot represent interactions between decision variables well. As a consequence, EDAs using univariate frequencies cannot efficiently solve complex problems. Using multivariate models allows complex problems to be solved, but fitting the model to a population and sampling new solutions can be very time-consuming. 

Recent work has shown that current models from machine learning such as the Restricted Boltzmann Machine (RBM), a stochastic neural network, can be used as probabilistic model for an EDA \cite{Probst2014}. While not entirely matching the quality of the more statistics-driven Bayesian Optimization Algorithm (BOA, \cite{Pelikan1999}), they have other desirable properties: speed of training and sampling, and easy and efficient parallelization \cite{Probst2014,Probst2014a}.

We focus on another model from the field of machine learning, which is closely related to the RBM - the Autoencoder (AE, see e.g.\cite{hinton2006reducing,Bengio2009deep}). Recent work has shown that AEs implicitly capture the probability distribution of given data, and that sampling this distribution is possible \cite{Bengio-et-al-NIPS2013}. Although the AE is structurally similar to an RBM, the training procedure is simpler and computationally less expensive. Hence, they are even faster to train and sample.

In this paper, we integrate a DAE in an EDA and assess its performance on multiple standard benchmark problems from combinatorial optimization. We report both the number of fitness evaluations and the required CPU times. We include results for BOA, RBM-EDA, and a simple univariate method for comparison.

Section \ref{preliminaries} introduces EDAs, Autoencoders, shows how to use an AE within an EDA and briefly discusses a related approach.  In Section \ref{experiments}, we present test problems, reference algorithms, experimental setup, and  results. We discuss the results in Section \ref{discussion} and conclude the paper in Section \ref{conclusion}.%
%==============================================================================%
%------------------------------------- PRELIMINARIES --------------------------%
%==============================================================================%
\section{Preliminaries}
\label{preliminaries}
We review the basic concept of EDAs. We introduce Autoencoders, describe how to train and sample them, and show how an Autoencoder can be used in an EDA.
%=========================== EDAs ============================
\subsection{Estimation of Distribution Algorithms}
EDAs are well-established tools for solving combinatorial optimization problems (see e.g. \cite{Muehlenbein1996,larranaga2002estimation}). The basic structure of EDAs is given by Algorithm \ref{alg-eda}. In a nutshell, they select promising individuals from a population, build a probabilistic model of this subpopulation and then use this model to sample new individuals. These new individuals are evaluated and usually form the new population. This loop continues until the population has converged. The underlying assumption is that a model, which has captured the essence of the old population, is able to sample new, unknown individuals that possess the same high-quality structure, thereby searching the solution space efficiently.

EDAs differ in their choice of the model. Simple models use a vector with activation probabilities for each variable of the problem, while neglecting dependencies between the variables, like UMDA or PBIL \cite{Muehlenbein1996,baluja1994population}. Slightly more complex models use pairwise dependencies modeled as trees or forests \cite{pelikan1999bivariate}. More complex dependencies can be captured by models with multivariate interactions, like ECGA or BOA \cite{harik2006linkage,Pelikan1999}.   Multivariate models are better suited for complex optimization problems, as univariate models can cause an exponential growth of the required number of fitness evaluations for growing problem sizes \cite{Pelikan1999, Pelikan2005}. Many algorithms use probabilistic graphical models with directed edges, i.e., Bayesian networks, or undirected edges, i.e., Markov random fields \cite{larranaga2012review}. Hence, model building consists of finding a network structure that matches the problem structure and estimating the model's parameters. Usually, the computational effort to build the model rises with model complexity and representational power.
\begin{algorithm}[t]
\caption{Pseudo code for main EDA loop}
\label{alg-eda}
\begin{algorithmic}[1]
\STATE \textbf{Initialize} Population $P$
\WHILE {not converged}
\STATE    $P_{parents}$ $\leftarrow$ \textbf{Select} high-quality solutions from $P$ based on their fitness
\STATE    $M$ $\leftarrow$ \textbf {Build} a model  estimating the  (joint) probability distribution of $P_{parents}$ 
\STATE    $P_{candidates}$ $\leftarrow$ \textbf{Sample} new candidate solutions from $M$
\STATE    $P$ $\leftarrow$ $P_ {parents}\cup P_ {candidates}$
\ENDWHILE
\end{algorithmic}
\end{algorithm}
%=========================== Autoencoders ============================
\subsection{Autoencoders}
This section shows how to train an AE, introduces the Denoising AE, and shows how to sample new solutions.
\subsubsection{Structure and Training Procedure}
\begin{figure}
\begin{center}
\centerline{\includegraphics[width=0.37\columnwidth]{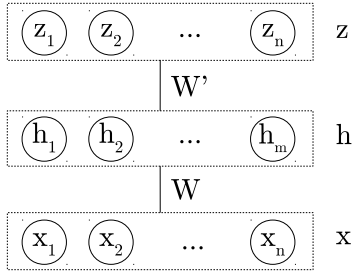}}
\caption{An Autoencoder as a graph. The visible neurons $x_i$ ($i\in{1..n}$) can hold a data vector of length $n$ from the training data. In the EDA context, each $x_i$ represents a decision variable. The hidden neurons $h_j$ ($j \in{1..m}$) form a compressed representation of the input. The output neurons $z_i$ ($i\in{1..n}$) hold the AE's \textit{reconstruction} of the input. Weights $W$ and $W'$ fully connect  $x$ to $h$ and $h$ to $z$ , respectively.}
\label{fig-ae}
\end{center}%
\end{figure}%
AEs are neural networks that have often been used for dimensionality reduction and are one of the building blocks for deep learning (see e.g. \cite{hinton2006reducing,Bengio2006Greedy,Bengio2009deep}). They are, in essence, multi layer perceptrons, which is a very basic type of neural network (see e.g. \cite{murphy2012machine}). 

An AE's structure is defined by one visible layer $x$, at least one hidden layer $h$, and one output layer $z$ (see Figure \ref{fig-ae}). The basic AE consists of two deterministic functions: the encoding function $h=c(x;\theta)$ maps a given input, $x\in[0,1]^n$, to a hidden layer, $h\in[0,1]^{m}$, with parameters $\theta$ and $n,m\in\mathbb{N}$. The decoding function $z=f(h;\theta')$, maps $h$ back to a \textit{reconstruction}  $z\in[0,1]^n$ in the input space. The training objective of the AE is to find parameters $\theta,\theta'$ which minimize the \textit{reconstruction error} $Err(x,z)$, i.e., the difference between $x$ and $z$ for all examples $x^i, i\in(1,\dots,\tau)$ in the training set:
\begin{equation}
\label{ae-objective}
\theta,\theta':=\underset{{\theta,\theta'}}{\text{argmin}} \frac{1}{\tau}\sum_{i=1}^{\tau}{\text{Err}(x^i,z^i)}.
\end{equation}
Common choices for $\text{Err}(x,z)$ are the mean squared error function $\text{Err}(x,z)=||x-z||^2$ or the cross entropy function $\text{Err}(x,z)=-\sum_{k=1}^{n}[x_k*\log(z_k) + (1-x_k)*\log(1-z_k)]$.

Encoding and decoding functions are usually chosen as $c(x)=\text{sigm}({x*W+b_h})$ and $f(h)=\text{sigm}({h*W'+b_z})$, where $\text{sigm}(x)=\frac{1}{1+e^{-x}}$ is the logistic function, $W$ and $W'$ are weight matrices of size $(n\times m)$ and $(m\times n)$, respectively, and $b_h\in{\mathbb{R}^{m}}$, $b_z\in{\mathbb{R}^{n}}$ are biases which work as offsets. Often, $W$ and $W'$ are \textit{tied}, i.e., $W'=W^\top$. Then, the AEs configurable parameters are $\theta=\{W,b_h,b_z\}$.

Minimizing (\ref{ae-objective}) is performed by using a gradient descent algorithm (see Algorithm \ref{alg-ae}). First, the parameters $\theta$ are initialized to small, random values. Then, we repeat the following process for multiple epochs, i.e., passes through the training set: For each example $x^i$ in the training set we calculate the hidden layer, $h=c(x^i;\theta)$, and the corresponding reconstruction $z=f(h;\theta)$. We then change the parameters in the direction of the gradient, setting
\begin{equation}
\label{ae-update}
\theta:=\theta-\alpha*\frac{\partial\text{Err}(x,z)}{\partial\theta},
\end{equation}
 with learning rate $0<\alpha<1$. We stop the loop if the reconstruction error is small enough or another termination criterion has been met.
Often, the parameter optimization is carried out using \textit{stochastic} gradient descent, i.e., we use the average gradient from a \textit{mini-batch} of $b$ training examples to update $\theta$. This usually speeds up learning and makes the gradient more stable \cite{bishop2006pattern}.

\begin{algorithm}[t]
\caption{Pseudo code for training an AE}
\label{alg-ae}
\begin{algorithmic}[1]
\STATE \textbf{Initialize} $\theta=\{W,b_h,b_z\}$ randomly
\STATE \textbf{Set} $0<\alpha<1$, e.g. $\alpha=0.1$
\WHILE {not converged}
\FOR{\textbf{each} example $i$ in the training set}
\STATE $h=c(x^i;\theta)$
\STATE $z=f(h;\theta)$
\STATE $\theta:=\theta-\alpha*\frac{\partial\text{Err}(x^i,z)}{\partial\theta}$
\ENDFOR
\ENDWHILE
\STATE (for training a DAE, replace $x^i$ with $q(\hat{x}^i|x^i)$ in line 5)
\end{algorithmic}
\end{algorithm}
\subsubsection{Denoising AE}
If the representational power of the hidden layer $h$ is large enough (i.e., if $m$ is not too small), a trivial way to solve (1) is to learn the identity function where each $x_i$ is directly mapped to the corresponding $z_i$ \cite{alain2014regularized}. To force the model to learn a more useful representation, it is therefore often helpful to introduce a form of regularization \cite{Bengio2009deep,alain2014regularized}. One example of a regularized AE is the Denoising Autoencoder (DAE) introduced in \cite{vincent2008extracting}. Here, each training example $x$ is corrupted by a stochastic mapping $\hat{x}=q(\hat{x}|x)$, i.e., we add random noise. Subsequently, the DAE calculates the reconstruction of the corrupted input, using encoding and decoding function, as $z=f(c(\hat{x}))$. As with the original AE, the parameters are updated in the direction of $\frac{\partial\text{Err}(x,z)}{\partial\theta}$. Hence, the DAE tries to reconstruct $x$ rather than $\hat{x}$.
The noise introduced by the corruption process $q(\cdot)$ also makes the model more robust to partially destroyed inputs \cite{vincent2008extracting}.
\subsubsection{Sampling a DAE}
\label{dae:sampling}
Classic AEs do not include a sampling process to generate new examples. However, recent work has shown that some variants of AEs, including the DAE, implicitly capture the structure of the data-generating density, and multiple sampling processes have been suggested and empirically validated (for an overview, see \cite{Bengio-et-al-NIPS2013}).
Here, we adopt the sampling process proposed in \cite{Bengio-et-al-NIPS2013}, because it is the most general approach, and comes with a theoretical justification.

Given a data-generating distribution, $P(x)$, a corruption process, $q(\hat{x}|x)$ and a DAE that has been trained to reconstruct $x$ from $\hat{x}$, the sampling process is as follows (see Algorithm \ref{alg-dae-sampling}): First, we randomly initialize a sample $x\in[0,1]^n$. Then, for $s$ sampling steps, we corrupt the sample using the corruption process $\hat{x}=q(\hat{x}|x)$ and use the trained DAE to reconstruct the input $z=f(c(\hat{x}))$. For the next sampling step, we set $x:=z$. After $s$ sampling steps, we use $x$ as a sample from the DAE.

In \cite{Bengio-et-al-NIPS2013}, it was shown that this process converges to samples from the DAE's approximation of the data-generating distribution, i.e., the training data.
\begin{algorithm}[t]
\caption{Pseudo code for sampling a DAE}
\label{alg-dae-sampling}
\begin{algorithmic}[1]
\STATE \textbf{Given} the trained DAE's $\theta=\{W,b_c,b_f\}$ and its reconstruction function $f(c(\hat{x}))$, the corruption process $q(\hat{x}|x)$
\STATE \textbf{Initialize} $x\in[0,1]^n$ randomly
\FOR{\textbf{a fixed number $s$ of sampling steps}}
\STATE $\hat{x}=q(\hat{x}|x)$
\STATE $z=f(c(\hat{x}))$
\STATE $x:=z$
\ENDFOR
\STATE Use $x$ as a sample from the DAE
\end{algorithmic}
\end{algorithm}

\subsection{Using a Denoising Autoencoder in an EDA}
We can use a DAE as probabilistic model for an EDA. In each generation of the EDA, we train a DAE to model the probability distribution of the solutions which survived the selection process. We then sample the DAE. Each sample is a vector $x\in[0,1]^n$. To turn this vector of real-valued elements into a candidate solution, i.e., a binary string, we sample each variable $x_i$ from a Bernoulli distribution with $p=x_i$. Then, we evaluate the fitness of the candidate solutions, and let the selection function decide which individuals will reach the next generation.

Another approach for using a DAE in an EDA-like optimization process was recently suggested by \cite{Churchill2014}. Contrary to our approach, the DAE in \cite{Churchill2014} is not used as a multivariate EDA model to sample new solutions. Instead, it is trained on the best 10-20\% of the population only. Subsequently, it is used to improve a second set of selected individuals from the population. Those individuals are first corrupted by the DAE's corruption process, and then reconstructed by the DAE, using encoding and decoding function.
%==============================================================================
%------------------------------------- EXPERIMENTS ----------------------------
%==============================================================================
\section{Experiments}
\label{experiments}
We present test problems, reference algorithms, experimental setup, and results.
%=========================== Test Problems ============================
\subsection{Test Problems}
\label{test_problems}
We evaluate DAE-EDA on concatenated deceptive traps, NK landscapes and the HIFF function. All three are standard benchmark problems. Their difficulty depends on the problem size, i.e., problems with more decision variables are more difficult. Furthermore, the difficulty of concatenated deceptive trap functions and NK landscapes is tunable by a parameter. All three problems are composed of subproblems, which are either deceptive (traps), overlapping (NK landscapes), or hierarchical (HIFF), and therefore multimodal. 

Concatenated deceptive traps are tunably hard, yet decomposable test problems \cite{deb1993analyzing}. Here, a solution vector $x$ is divided into $l$ subsets of size $k$, with each one being a deceptive trap. Within a trap, all bits are dependent on each other but independent of all other bits in $x$. Thus, the fitness contribution of the traps can be evaluated separately and the total fitness of the solution vector is the sum of these terms.
In particular, the assignment $a=x_{i:i+k-1}$  (i.e., the $k$ bits from $x_i$ to $x_{i+k-1})$\footnote{The $k$ variables assigned to trap $l$ do not have to be adjacent, but can be at any position in $x$.} leads to a fitness contribution $F_l$ as
$$
F_l(a) = \begin{cases} k &\mbox{if } \sum_{i}{a_i}=k, \\
k-(\sum_{i}{a_i}+1) & \mbox{otherwise.}
\end{cases}
$$
In other words, the fitness of a single trap increases with the number of zeros, except for the optimum of all ones.

NK landscapes are defined by two parameters $n$ and $k$ and $n$ fitness components $f_{i}, i\in\{1\,\dots,n\}$ \cite{kauffman1989nk}. A solution vector $x$ consists of $n$ bits. The bits are assigned to $n$ overlapping subsets, each of size $k+1$. The fitness of a solution is the sum of $n$ fitness components.
Each component $f_{i}$ depends on the value of the corresponding variable $x_i$ as well as $k$ other variables. Each $f_{i}$ maps each possible configurations of its $k+1$ variables to a fitness value. The overall fitness function is
$$
f(x)=1/n\sum_{i=1}^nf_i(x_i,x_{i1},\ldots,x_{iK}).
$$ 
\noindent Each decision variable usually influences several $f_i$. These dependencies between subsets make NK landscapes non-separable. The problem difficulty increases with $k$. $k=0$ is a special case where all decision variables are independent and the problem reduces to a unimodal onemax. We use instances of NK landscapes with known optima from \cite{Pelikan2008techreport}.

The  Hierarchical If-and-only-if (HIFF) function \cite{watson1998modeling} is defined for solutions vectors of length $n=2^l$ where $l\in\mathbb{N}$ is the number of layers of the hierarchy. It uses a mapping function $M$ and a contribution function $C$, both of which take two inputs.
The mapping function takes each of the $n/2$ blocks of two neighboring variables of level $l=1$, and maps them onto a single symbol each. An assignment of $00$ is mapped to $0$, $11$ is mapped to $1$ and everything else is mapped to the null symbol '-'. The concatenation of $M$'s outputs on level $l$ is used as M's input for the next level $l+1$ of the hierarchy, i.e., if level $l=1$ has $n$ variables, level $l=2$ has $n/2$ variables.
On each level, $C$ assigns a fitness to each block of two variables. The assignments $00$ and $11$ are both mapped to $2^l$, everything else to $0$. The total fitness is the sum of all blocks' contributions on all levels.
In other words, a block contributes to the fitness on the current level if both variables in a block have the same assignment. However, only if neighboring blocks agree on the assignment, they will contribute to the fitness on the next level. HIFF therefore has two global optima, the string of all ones, and the string of all zeros.
%=========================== Reference Algorithms ============================
\subsection{Reference Algorithms}
\label{reference Algorithms}
We compare DAE-EDA to BOA \cite{Pelikan1999}, RBM-EDA, an EDA based on Restricted Boltzmann Machines \cite{Probst2014}, and Population-Based Incremental Learning (PBIL, \cite{baluja1994population}).
\subsubsection{Bayesian Optimization Algorithm}
\label{boa}
The Bayesian Optimization Algorithm is one of the state-of-the-art EDAs for discrete optimization problems. It was been proposed by \cite{Pelikan1999} and has been heavily used and researched since then \cite{pelikan2003hierarchical,Pelikan2008techreport,abdollahzadeh2012bayesian}. 

BOA uses a Bayesian network for modeling dependencies between variables. Decision variables correspond to nodes and dependencies between variables correspond to directed edges. As the number of possible network topologies grows exponentially with the number of nodes, BOA uses a greedy construction heuristic to find a network structure $G$ to model the training data. Starting from an unconnected (empty) network, BOA evaluates all possible additional edges, adds the one that maximally increases the fit between the model and selected individuals, and repeats this process until no more edges can be added. The fit between selected individuals and the model is measured by the Bayesian Information Criterion (BIC) \cite{schwarz1978estimating}. BIC is based on the conditional entropy of nodes given their parent nodes and correction terms penalizing complex models. It can be calculated independently for all nodes. If an edge is added to the Bayesian network, the change of the BIC can be computed quickly. BOAs greedy network construction algorithm adds the edge with the largest BIC gain until no more edges can be added. Edge additions resulting in cycles are not considered. 

After the network structure has been learned, BOA calculates the conditional activation probability tables for each node. Once the model structure and conditional activation probabilities are available, BOA can produce new candidate solutions by drawing random values for all nodes in topological order.
\subsubsection{RBM-EDA}
RBM-EDA uses a Restricted Boltzmann Machine as multivariate model. Restricted Boltzmann Machines are stochastic neural networks consisting of two layers of neurons, where the connections between the layers form a bipartite graph \cite{Smolensky1986}.
The input or visible layer $x\in[0,1]^n$ of an RBM holds the input data represented by $n$ binary variables.  The second, hidden layer $h\in[0,1]^m$ consists of $m$ neurons. There is no dedicated output layer in an RBM. A weight matrix $W$ holds weights $w_{i,j}\in \mathbb{R}$ between all neurons $x_i$ and $h_j$. From a structural point of view, an RBM resembles an Autoencoder with one hidden layer where the output layer has been "folded" back onto the input layer.\\
An RBM can be used as a model within an EDA, because it can be trained to model a probability distribution and it is possible to draw samples from this model \cite{Smolensky1986,Hinton2002,Hinton2006}. Training the RBM means adjusting $w_{i,j}$ s.t. the RBM models the probability distribution of the training data. This can be done by using the gradient descent algorithm \textit{contrastive divergence} \cite{Hinton2002}. Sampling new individuals from the model's probability distribution can be performed using Gibbs sampling \cite{GemanGeman1993}.

\cite{Probst2014} have shown that RBM-EDA is competitive to BOA. For difficult problems, it has a moderately higher, but still non-exponential complexity in the number of fitness evaluations. However, the time for solving problems grows slower with larger problem sizes. We compare DAE-EDA to RBM-EDA, because they are closely related in terms of the models' structure and training process.
\subsubsection{PBIL}
PBIL is one of the simplest EDAs. It assumes conditional independence of all $n$ problem variables. PBIL stores a vector $P=(p_1,p_2,\dots,p_n)$ of activation probabilities. PBIL creates new individuals by sampling each variable from a Bernoulli distribution with $p=p_n$. In each EDA generation $t$, PBIL selects the best $\mu$ individuals $y_1,y_2,\dots,y_k$ from the population, and updates each $p_n$ as 
$$
p_n^{t+1}=p_n^{t}+\alpha*[(\frac{1}{\mu}\sum_{k=1}^{\mu}{y_k^t})-p_n^t],
$$
with $0<\alpha<1$ determining the strength of the update. We include PBIL in the experiment, because its results give an intuitive measure on the difficulty of the test problems.
%=========================== Experimental setup ============================
\subsection{Experimental Setup}
\label{experimental_setup}
\subsubsection{EDA Parametrization}
We use several instances of the test problems (see section \ref{results}). For each instance and algorithm, we test multiple population sizes. For DAE-EDA, RBM-EDA, and BOA, we choose $\text{popsize}\in\{50;100;\dots;16,000\}$, for PBIL, we choose $\text{popsize}\in\{50;100;\dots;512,000\}$). We run 20 instances for each population size.

In each run, the EDA is allowed to run for 100 generations (2000 for PBIL). We terminate the EDAs if there is no improvement in the best solution for more than 20 generations (400 for PBIL). We report the average number of fitness evaluations and CPU time for the best solutions of all runs.  All EDAs use tournament selection without replacement of size two \cite{Miller95geneticalgorithms}.

For PBIL, we choose $\mu=1$ and $\alpha=.02$., i.e., we use only the best individual in each generation to update the model. For the RBM, we use the same parameter settings as in \cite{Probst2014}. 

The algorithms were implemented in Matlab/Octave and executed using Octave V3.2.4 on a on a single core of an AMD Opteron 6272 processor with 2,100 MHz.  
\subsubsection{DAE Parametrization}
We use the following parameters for the DAE: The number $m$ of hidden neurons is equal to the problem size $n$. The corruption process $q(\hat{x}|x)$ randomly corrupts 10\% of the inputs by setting them to 0 or 1 (salt+pepper noise). When sampling new candidate solutions from the DAE, we perform $s=10$ sampling steps. During training, the learning rate $\alpha$ is 0.2, the batch size for stochastic gradient descent is $b=100$. We use a cross-entropy error measure for (\ref{ae-objective}).

Like \cite{Probst2014}, we apply a simple parameter control scheme determining when to terminate DAE training.
The scheme is based on the reconstruction error $e=Err(x,z)$. $e$ usually decreases with the number of epochs. Every second epoch $t\in{1,\ldots,T}$, we calculate for a fixed subset $u$ of the training set $U$ the relative difference $e_t^u=1/{|u|}\sum_{j\in u}Err(x^j,z^j).$
We measure the decrease $\gamma$ of the reconstruction error in the last 33\% of all epochs as
$\gamma=(e^u_{0.67t}-e^u_t)/(e^u_0-e^u_t).$
$\gamma$ is then used to automatically check for convergence of the training. We stop training if $\gamma<0.05$. The rationale behind this is that the DAE has learned the relevant dependencies between the variables, and further training is unlikely to improve the model considerably.
Furthermore, we stop the training if the DAE is overfitting, i.e., learning noise instead of problem structure. Therefore, we split the original training set into a training set $U$ containing 90\% of all samples and a validation set $U'$ containing the remaining 10\%. We train the DAE only for the solutions in $U$ and, after each epoch, calculate the reconstruction error $e^{U}$ and $e^{U'}$ for the training and validation set $U$ and $U'$, respectively. We stop the training phase as soon as $(|e^{u}-e^{U'}|)/e^{u}\geq 0.1$ (i.e., the difference between the reconstruction errors is larger than 10\%). 

%=========================== Results ============================
\subsection{Results}
\label{results}
We report the performance of DAE-EDA, RBM-EDA, BOA, and PBIL for concatenated deceptive traps with $k=4,n\in\{20,40,60\}$ and $k=5,n\in\{25,50,75\}$, NK landscapes with $k=4,n\in\{30,34\}$ and $k=5,n\in\{30,34\}$ (two instances $i$ each) as well as the HIFF function with $n\in\{64,128\}$ (see table \ref{table:results}).
For each instance and algorithm, we select the minimal population size which leads to the optimal solution in at least 50\% of the runs (left two result columns of table \ref{table:results}) and at least 90\% of the runs (right two columns). We report the average number of fitness evaluations and CPU time of those runs. 

First, we analyze the number of fitness evaluations required.
For all problems, and for both the runs with at least 50\% and 90\% success rate, respectively, PBIL uses the highest number of fitness evaluations. 

As expected, BOA has the best performance in terms of fitness evaluations. This is consistent with the previous findings comparing RBM-EDA and BOA \cite{Probst2014}.

Both DAE-EDA and RBM-EDA consistently use more fitness evaluations than BOA, except for the NK landscapes with $n=34,k=5$. However, most of the time the number of fitness evaluations is on the same order of magnitude, and clearly better than that of the univariate PBIL. For the both the runs with at least 50\% and 90\% success rate, DAE-EDA and RBM-EDA are about tied for the number of instances with the least fitness evaluations.  

For the 128 bit HIFF problem DAE-EDA needed only 45\% of the fitness evaluations of the DAE inspired optimizer in \cite{Churchill2014}. We attribute this mainly to the sampling process, which samples from the trained model's distribution directly, instead of using the DAE as a tool for an advanced local search modifying selected individuals.

Second, we look at the average time the algorithms required to solve the respective problem. If PBIL is able to solve the problem to optimality, it is usually the fastest algorithm, because fitness evaluations are computationally inexpensive for all benchmark problems. Note that for the 60-bit concatenated  4-trap problem, PBIL is able to find the optimal solution in at least 50\% of the runs, nevertheless, the DAE-EDA is faster as it needs only about 1\permil ~of PBIL's fitness evaluations.

For all but one instance, DAE-EDA is significantly faster than both RBM-EDA and BOA, sometimes by multiple orders of magnitude (see Section \ref{discussion}).
This is also true for the instances where RBM-EDA needs a lower number of fitness evaluations. This is due to the much quicker model building of the DAE.
\begin{table*}
\scriptsize
\renewcommand{\arraystretch}{1.0}
%%%%%% BEGIN TABLE FROM SCRIPT, do not change manually 
\begin{tabular}{| c | c ||r@{\hskip 0cm} l|r@{\hskip 0cm} l|r@{\hskip 0cm} l|r@{\hskip 0cm} l|}
\hline
\multirow{2}{1.7cm}{\textbf{Problem}}&\multirow{2}{1.7cm}{\textbf{Algorithm}}&\multicolumn{8}{c|}{\shortstack{\textbf{Average results}\\Population size such that optimum is found}}
\\

&&\multicolumn{4}{c|}{in $\geq$50\% of runs}
&\multicolumn{4}{c|}{in $\geq$90\% of runs}
\\

&&\multicolumn{2}{c|}{Evaluations}&\multicolumn{2}{c|}{Time (sec)}&\multicolumn{2}{c|}{Evaluations}&\multicolumn{2}{c|}{Time (sec)}\\
\hline 
\multirow{4}{1.7cm}{4-Traps 20~bit}&DAE-EDA&2,550&$\pm$1,150&18&$\pm$6.9&4,450&$\pm$1,359&21&$\pm$9.2\\
&RBM-EDA&20,300&$\pm$7,753&121&$\pm$30&20,300&$\pm$7,753&121&$\pm$30\\
&BOA&\textbf{1,850*}&$\pm$382&117&$\pm$26&\textbf{1,850*}&$\pm$382&117&$\pm$26\\
&PBIL&23,050&$\pm$8,968&\textbf{0.0*}&$\pm$0.3&71,200&$\pm$31,688&\textbf{0.0*}&$\pm$0.4\\
\hline 
\multirow{4}{1.7cm}{4-Traps 40~bit}&DAE-EDA&37,400&$\pm$14,030&90&$\pm$28&37,400&$\pm$14,030&90&$\pm$28\\
&RBM-EDA&45,000&$\pm$14,346&696&$\pm$144&56,000&$\pm$9,633&656&$\pm$72\\
&BOA&\textbf{7,875*}&$\pm$1,035&1,963&$\pm$304&\textbf{7,875*}&$\pm$1,035&1,963&$\pm$304\\
&PBIL&3,398,400&$\pm$1,276,058&\textbf{38*}&$\pm$14&6,121,600&$\pm$2,199,336&79&$\pm$27\\
\hline 
\multirow{4}{1.7cm}{4-Traps 60~bit}&DAE-EDA&61,800&$\pm$22,225&\textbf{182*}&$\pm$58&292,000&$\pm$55,857&\textbf{823*}&$\pm$216\\
&RBM-EDA&95,000&$\pm$16,823&1,822&$\pm$192&163,600&$\pm$28,675&1,842&$\pm$268\\
&BOA&\textbf{18,600*}&$\pm$1,655&10,658&$\pm$2,321&\textbf{18,600*}&$\pm$1,655&10,658&$\pm$2,321\\
&PBIL&71,884,800&$\pm$19,416,482&1,229&$\pm$380&-&&-&\\
\hline 
\multirow{4}{1.7cm}{5-Traps 25~bit}&DAE-EDA&11,650&$\pm$5,350&44&$\pm$13&11,650&$\pm$5,350&44&$\pm$13\\
&RBM-EDA&34,600&$\pm$8,511&230&$\pm$40&48,800&$\pm$14,288&231&$\pm$35\\
&BOA&9,550&$\pm$2,061&1,099&$\pm$299&13,000&$\pm$2,049&1,572&$\pm$298\\
&PBIL&180,000&$\pm$63,182&\textbf{1.0*}&$\pm$0.6&564,000&$\pm$265,083&\textbf{4.0*}&$\pm$2.0\\
\hline 
\multirow{4}{1.7cm}{5-Traps 50~bit}&DAE-EDA&57,750&$\pm$18,250&\textbf{200*}&$\pm$44&57,750&$\pm$18,250&\textbf{200*}&$\pm$44\\
&RBM-EDA&89,000&$\pm$18,615&841&$\pm$180&119,000&$\pm$10,440&812&$\pm$126\\
&BOA&\textbf{34,600*}&$\pm$2,615&15,905&$\pm$2,218&43,800&$\pm$3,341&20,196&$\pm$2,552\\
&PBIL&-&&-&&-&&-&\\
\hline 
\multirow{4}{1.7cm}{5-Traps 75~bit}&DAE-EDA&96,500&$\pm$32,049&\textbf{519*}&$\pm$111&247,500&$\pm$45,373&\textbf{1,297*}&$\pm$233\\
&RBM-EDA&218,000&$\pm$32,496&2,235&$\pm$219&218,000&$\pm$32,496&2,235&$\pm$219\\
&BOA&82,000&$\pm$7,642&88,599&$\pm$11,852&\textbf{132,400*}&$\pm$9,952&145,026&$\pm$18,865\\
&PBIL&-&&-&&-&&-&\\
\hline 
\multirow{4}{1.7cm}{NK $n=30$, $k=4$, $i=1$}&DAE-EDA&10,725&$\pm$3,976&52&$\pm$12&31,900&$\pm$6,971&\textbf{79*}&$\pm$16\\
&RBM-EDA&47,300&$\pm$18,036&654&$\pm$135&55,000&$\pm$7,362&787&$\pm$133\\
&BOA&9,500&$\pm$1,975&1,185&$\pm$265&32,500&$\pm$5,723&4,161&$\pm$929\\
&PBIL&430,400&$\pm$227,594&\textbf{6.0*}&$\pm$3.6&-&&-&\\
\hline 
\multirow{4}{1.7cm}{NK $n=30$, $k=4$, $i=2$}&DAE-EDA&49,100&$\pm$13,423&109&$\pm$24&328,000&$\pm$77,974&\textbf{396*}&$\pm$108\\
&RBM-EDA&124,800&$\pm$22,400&1,133&$\pm$155&238,400&$\pm$40,128&1,490&$\pm$237\\
&BOA&45,500&$\pm$12,114&6,105&$\pm$1,741&\textbf{65,600*}&$\pm$11,056&12,854&$\pm$2,877\\
&PBIL&742,400&$\pm$292,734&\textbf{10*}&$\pm$4.3&-&&-&\\
\hline 
\multirow{4}{1.7cm}{NK $n=34$, $k=4$, $i=1$}&DAE-EDA&51,300&$\pm$16,844&126&$\pm$31&96,800&$\pm$24,351&171&$\pm$46\\
&RBM-EDA&48,800&$\pm$14,918&773&$\pm$222&63,000&$\pm$8,473&935&$\pm$132\\
&BOA&\textbf{11,900*}&$\pm$1,678&1,858&$\pm$413&\textbf{20,550*}&$\pm$3,556&3,296&$\pm$738\\
&PBIL&155,500&$\pm$51,170&\textbf{3.0*}&$\pm$1.2&5,088,000&$\pm$1,084,512&\textbf{84*}&$\pm$22\\
\hline 
\multirow{4}{1.7cm}{NK $n=34$, $k=4$, $i=2$}&DAE-EDA&\textbf{26,950*}&$\pm$8,176&99&$\pm$23&175,600&$\pm$45,218&\textbf{279*}&$\pm$81\\
&RBM-EDA&72,800&$\pm$15,677&1,168&$\pm$243&127,600&$\pm$13,260&1,441&$\pm$195\\
&BOA&42,700&$\pm$8,032&7,908&$\pm$1,908&\textbf{70,600*}&$\pm$9,902&16,500&$\pm$3,364\\
&PBIL&3,062,400&$\pm$1,126,159&\textbf{54*}&$\pm$20&-&&-&\\
\hline 
\multirow{4}{1.7cm}{NK $n=30$, $k=5$, $i=1$}&DAE-EDA&29,000&$\pm$8,349&98&$\pm$15&232,000&$\pm$56,114&323&$\pm$82\\
&RBM-EDA&\textbf{7,525*}&$\pm$1,847&56&$\pm$20&44,500&$\pm$7,124&634&$\pm$136\\
&BOA&11,825&$\pm$1,434&1,429&$\pm$283&\textbf{20,500*}&$\pm$3,892&2,742&$\pm$634\\
&PBIL&89,200&$\pm$40,355&\textbf{2.0*}&$\pm$1.2&4,006,400&$\pm$1,417,336&\textbf{68*}&$\pm$22\\
\hline 
\multirow{4}{1.7cm}{NK $n=30$, $k=5$, $i=2$}&DAE-EDA&99,600&$\pm$12,706&158&$\pm$25&182,400&$\pm$40,917&\textbf{240*}&$\pm$62\\
&RBM-EDA&46,700&$\pm$11,014&770&$\pm$153&69,200&$\pm$9,042&990&$\pm$144\\
&BOA&\textbf{22,150*}&$\pm$3,468&2,925&$\pm$595&\textbf{40,200*}&$\pm$9,421&5,492&$\pm$1,621\\
&PBIL&688,800&$\pm$263,708&\textbf{11*}&$\pm$4.1&-&&-&\\
\hline 
\multirow{4}{1.7cm}{NK $n=34$, $k=5$, $i=1$}&DAE-EDA&189,200&$\pm$63,964&\textbf{340*}&$\pm$122&309,600&$\pm$88,692&\textbf{455*}&$\pm$136\\
&RBM-EDA&\textbf{79,800*}&$\pm$12,867&1,235&$\pm$153&\textbf{149,200*}&$\pm$17,577&1,549&$\pm$207\\
&BOA&319,200&$\pm$44,535&97,558&$\pm$18,726&548,800&$\pm$84,830&178,667&$\pm$38,349\\
&PBIL&-&&-&&-&&-&\\
\hline 
\multirow{4}{1.7cm}{NK $n=34$, $k=5$, $i=2$}&DAE-EDA&213,200&$\pm$53,861&\textbf{347*}&$\pm$101&432,000&$\pm$92,330&\textbf{616*}&$\pm$143\\
&RBM-EDA&\textbf{138,400*}&$\pm$19,936&1,582&$\pm$255&\textbf{288,000*}&$\pm$31,190&2,240&$\pm$330\\
&BOA&280,000&$\pm$43,377&85,876&$\pm$15,708&436,800&$\pm$86,029&144,185&$\pm$33,991\\
&PBIL&-&&-&&-&&-&\\
\hline 
\multirow{4}{1.7cm}{HIFF64}&DAE-EDA&22,250&$\pm$4,097&\textbf{94*}&$\pm$15&36,900&$\pm$4,999&\textbf{141*}&$\pm$14\\
&RBM-EDA&43,200&$\pm$4,578&1,363&$\pm$122&75,800&$\pm$7,318&1,698&$\pm$174\\
&BOA&\textbf{11,175*}&$\pm$729&6,562&$\pm$1,068&\textbf{11,175*}&$\pm$729&6,562&$\pm$1,068\\
&PBIL&-&&-&&-&&-&\\
\hline 
\multirow{4}{1.7cm}{HIFF128}&DAE-EDA&103,400&$\pm$11,209&\textbf{870*}&$\pm$103&103,400&$\pm$11,209&\textbf{870*}&$\pm$103\\
&RBM-EDA&285,200&$\pm$19,964&9,543&$\pm$1,056&966,400&$\pm$85,026&25,462&$\pm$1,997\\
&BOA&\textbf{39,500*}&$\pm$3,122&101,067&$\pm$16,547&\textbf{69,300*}&$\pm$3,913&163,750&$\pm$22,829\\
&PBIL&-&&-&&-&&-&\\
\hline\end{tabular}
%%%%%%%%end table from script
\caption{\scriptsize This table shows average values for fitness evaluations and CPU time for DAE-EDA, RBM-EDA, BOA, and PBIL for the test problems. For each instance and algorithm, we selected the minimal population size which lead to the optimal solution in at least 50\% of the runs (left two result columns) and at least 90\% of the runs (right two columns). Results are averaged over 20 runs. Results marked with (*) are significantly smaller than other results in the respective table cell, according to pairwise Wilcoxon signed-rank tests ($p<0.01$, data is not normally distributed)}
\label{table:results}
\end{table*}
\section{Discussion}
\label{discussion}
The results suggest that DAE-EDA is able to decompose the test problems properly, and solve the parts independently. This becomes evident when looking at the more complicated problems where the univariate PBIL struggles or fails. The quality of DAE-EDAs underlying probabilistic model is similar to the one of RBM-EDA, but not as good as BOA's.

An interesting aspect of DAE-EDA is its speedy model building and sampling process. The CPU time to solve the test problems is much lower than that of the other multivariate methods, sometimes by multiple orders of magnitude. Note that the direct comparison of CPU times is not entirely fair for BOA. In a more efficient programming language instead of a script-based language like Matlab/Octave, BOA's speedup is significantly higher than the one of DAE-EDA and RBM-EDA. However, almost every recent implementation of neural networks is parallelized on graphics processing units (GPU), which, in turn, speeds up training and sampling these models considerably (see e.g. \cite{hinton2012improving,krizhevsky2012imagenet,Sutskever2014Sequence}). Parallelizing multivariate EDAs such as BOA is well possible, however the speedups are often single- or double-digit, even on GPUs (see e.g. \cite{ovcenavsek2000parallel,munawar2009theoretical}). In contrast, parallelizing EDAs using neural networks can saturate modern GPU hardware and yield very high speedups: \cite{Probst2014a} report speedups of up to 200$\times$, against optimized CPU code, for RBM-EDA, which uses a neural network model that is closely related to the DAE. Hence, it is reasonable to assume that an efficient GPU-based implementation of DAE-EDA will be very fast, compared to other EDAs on problems with low, but non-negligible fitness evaluation costs.

\begin{figure*}[t]
\centering
        \begin{subfigure}
                \centering
                \includegraphics[width=0.48\linewidth]{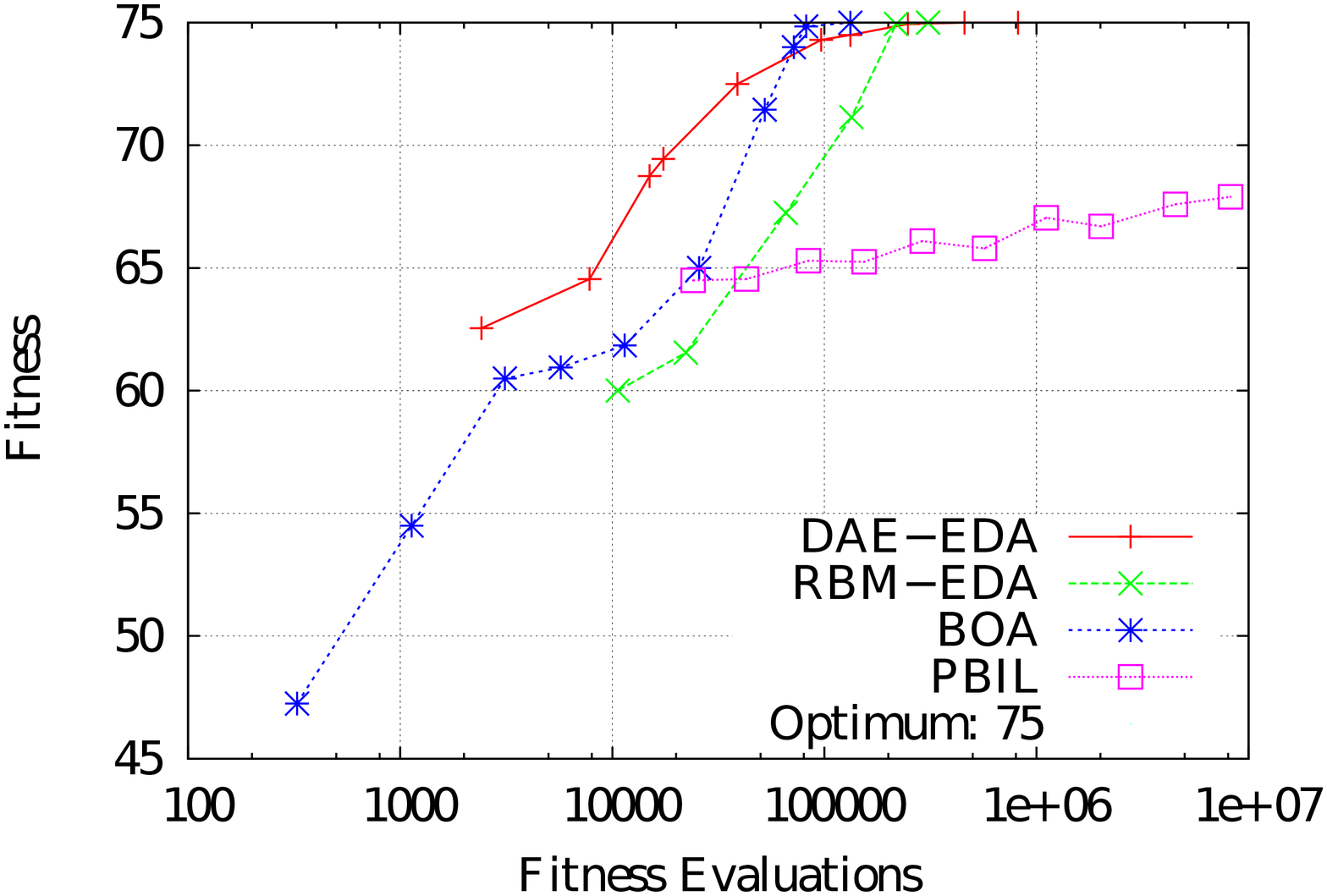}
                \label{fig-5T75-eval}
        \end{subfigure}
        \begin{subfigure}
                \centering
                \includegraphics[width=0.48\linewidth]{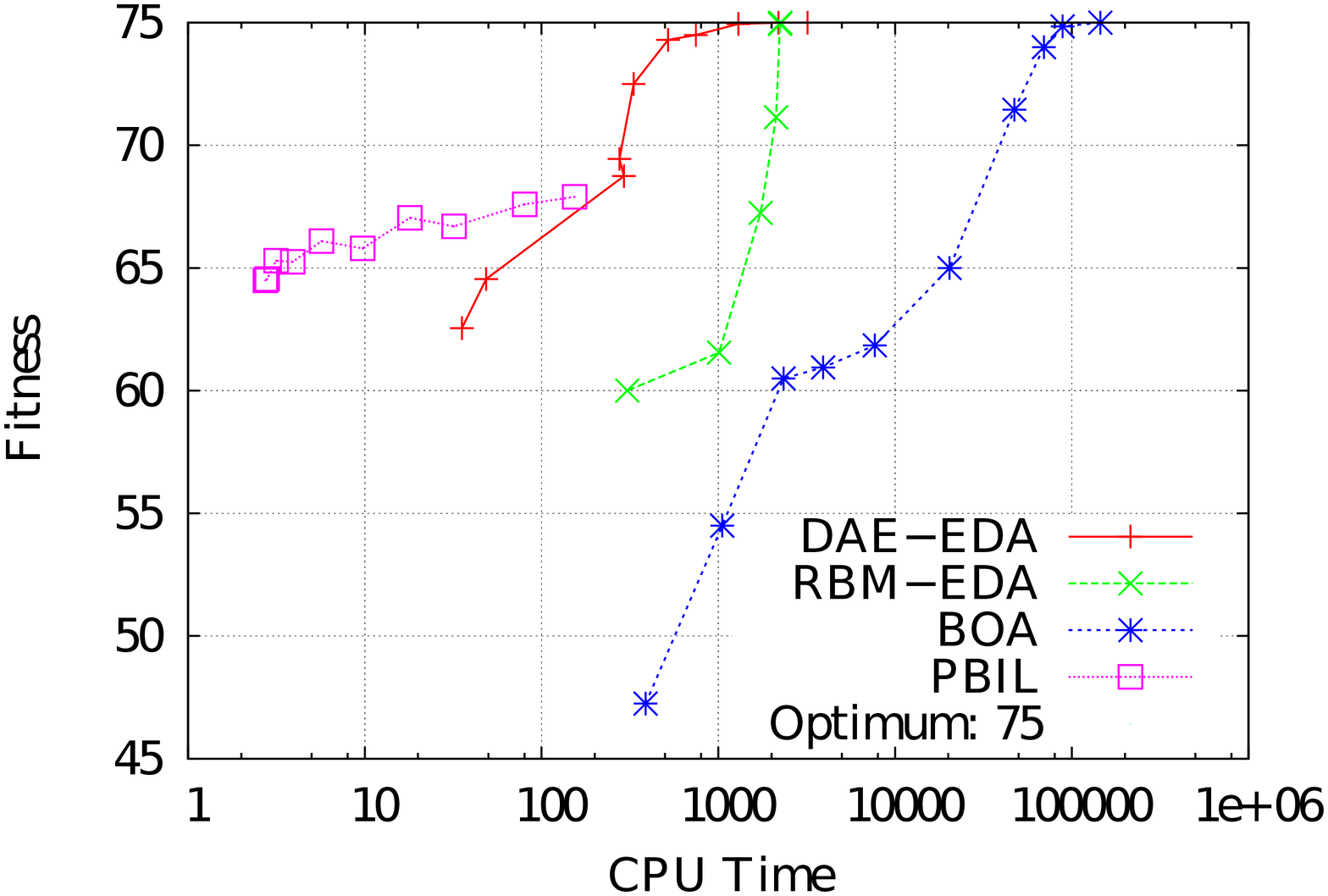}
                \label{fig-5T75-time}
        \end{subfigure}
        \caption{Number of fitness evaluations (left-hand side) and CPU time (right-hand side) for the 75 bit concatenated 5-Traps problem.}
        \label{fig-5T75}
\end{figure*}

\begin{figure*}[t]
\centering
        \begin{subfigure}
                \centering
                \includegraphics[width=0.48\linewidth]{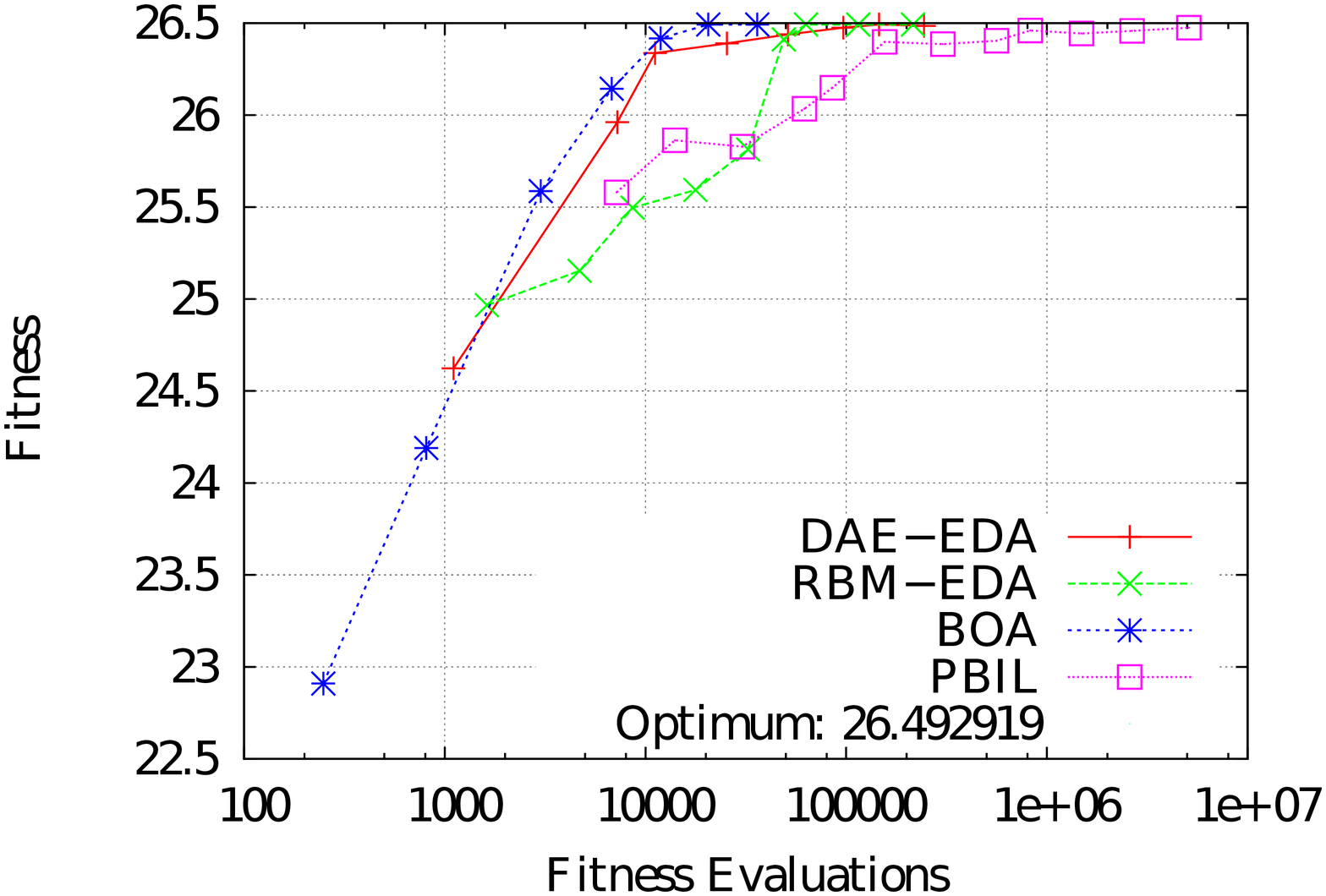}
                \label{fig-nk_nk_34_4.1-eval}
        \end{subfigure}
        \begin{subfigure}
                \centering
                \includegraphics[width=0.48\linewidth]{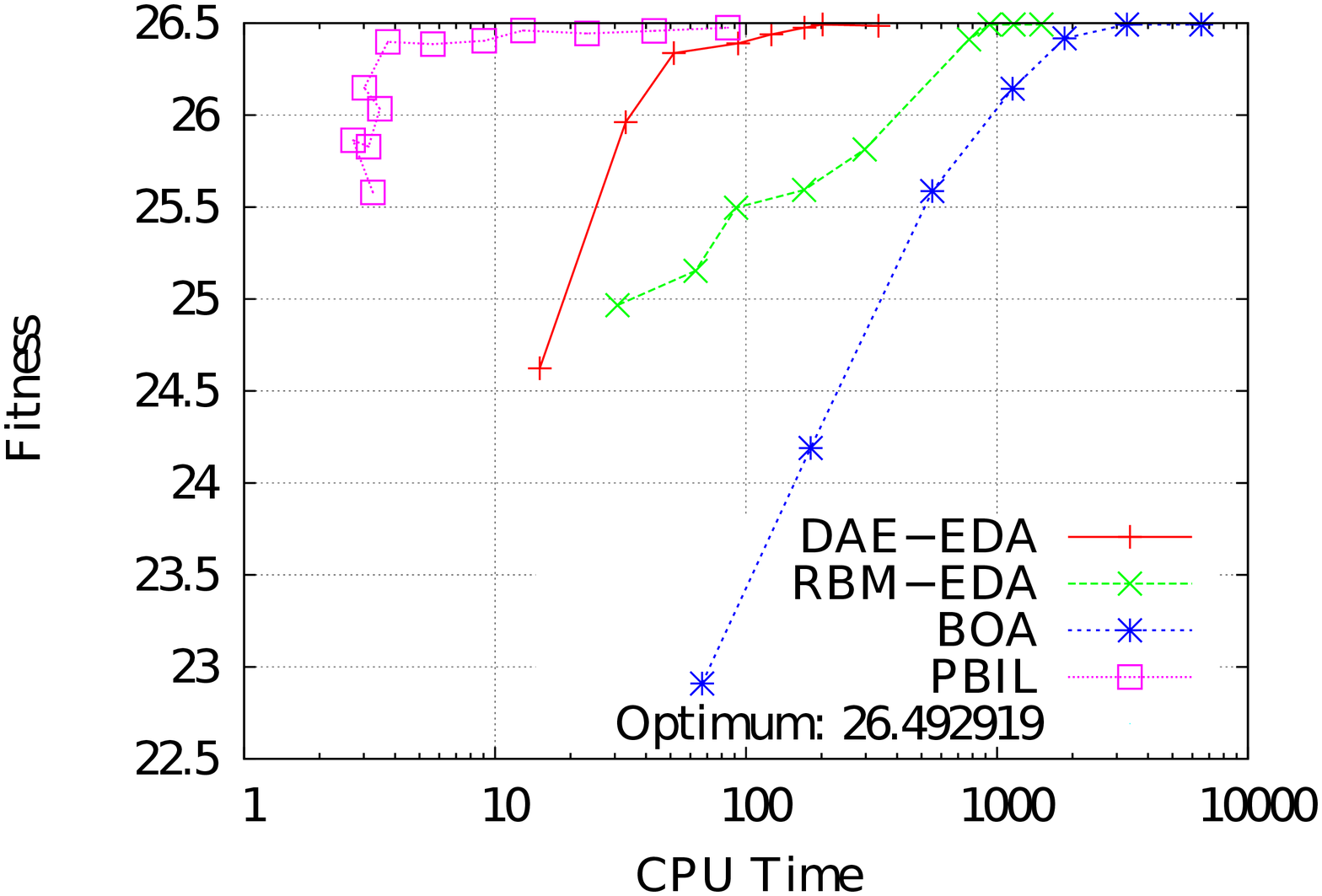}
                \label{fig-nk_34_4.1-time}
        \end{subfigure}
        \caption{Number of fitness evaluations (left-hand side) and CPU time (right-hand side) for an NK landscape with $n=34$,$k=4$ (instance 1).}
        \label{fig-nk_34_4.1}
\end{figure*}

Regarding the model quality, we now exemplarily take a closer look at the results of two selected test instances, where DAE-EDA has a higher number of fitness evaluations than both BOA and RBM-EDA: the 75 bit concatenated 5-trap problem (Figure \ref{fig-5T75}) and the NK landscape with $n=34$,$k=4,i=1$ (Figure \ref{fig-nk_34_4.1}). In both Figures, the left-hand side shows the number of fitness evaluations, the right-hand side shows the CPU time. Each data point of a problem marks the average results for a specific population size. Lines connect data points with adjacent population sizes.

For the 75 bit 5-trap problem, we see that DAE-EDA approaches the optimum faster (i.e., with smaller population size, fewer fitness evaluations and less total time) than both RBM-EDA and BOA. For the NK landscape, DAE-EDA is comparable to BOA.
However, for both problems, the transition region from partial success to complete success is larger for DAE-EDA. In other words, DAE-EDA quickly finds optimal or close-to-optimal solutions with few evaluations in some runs, but often needs much larger populations for \textit{reliable} convergence in every run.
This pattern is qualitatively similar for other problems: On average, DAE-EDA needs 2.5$\times$ the number of fitness evaluations to make $\geq$90\% rather than $\geq$50\% of runs converge, compared to 1.9$\times$ for RBM-EDA and $1.6\times$ for BOA (see table \ref{table:results}). 
This suggests that, in the current configuration, DAE-EDA is more dependent on the initialization of the DAE's parameters $\theta$. If, by chance, they are initialized in a particularly unfavorable way, the model is not able to learn an optimal hidden representation. The amount of random noise injected by corruption function $q(\cdot)$ is not sufficient to compensate for this effect. This hinders DAE-EDA from exploring the solution space more efficiently. This poses an interesting area for further research.%
%witdh max: 0.34
%==============================================================================%
%------------------------------------- CONCLUSION -----------------------------%
%==============================================================================%
\vspace{-0.1cm}
\section{Conclusion}
\label{conclusion}
We introduced DAE-EDA, an Estimation of Distribution Algorithm which uses a Denoising Autoencoder as probabilistic model for solving combinatorial optimization problems. DAE-EDA uses a DAE to approximate the probability distribution of the fittest individuals and subsequently samples new candidate solutions from this distribution. We tested DAE-EDA using several instances of  the standard benchmark problems concatenated deceptive traps, NK landscapes and the HIFF function. We compared the results to multiple other EDAs: The state-of-the-art Bayesian Optimization Algorithm, the multivariate RBM-EDA, another EDA based on stochastic neural networks, which has shown to be computationally less expensive than the state of the art BOA for complicated problems, and the univariate PBIL.

The results show that DAE-EDA is a very fast EDA that is able to decompose complicated problems. It needs a similar number of fitness evaluations like RBM-EDA, but does not reach the model quality of BOA. However, it is much faster than both RBM-EDA and BOA, as training and sampling the probabilistic model are conceptually simpler and computationally cheaper. Furthermore, a DAE is structurally similar to an RBM. Hence, we can assume the speedup of running a parallelized version of DAE-EDA on modern graphics processing units to be very high.

In sum, DAE-EDA can be a useful tool for solving complex combinatorial optimization problems, where fitness evaluation costs are low, but non-negligible.

There are multiple directions for further research. The results suggest that it could be beneficial to look into the parameter initialization more thoroughly, as DAE-EDA's performance seems to be more susceptible to unfavorable initial parameters.
Another promising direction is to use a deep, multi-layered DAE to solve hierarchical problems. Also, other techniques for sampling a DAE exist, which may result in a different performance of DAE-EDA. 

\section{Acknowledgments}
The authors would like to thank the anonymous reviewers for their valuable comments and suggestions on an earlier version of this paper.

\bibliographystyle{abbrv}
\small
\bibliography{references}

\end{document}